\documentclass[10pt]{article}
\usepackage[utf8]{inputenc}

\usepackage[left=1.25in, right=1.25in, top=1.25in, bottom=1.25in]{geometry}
\usepackage{graphicx}
\usepackage{xcolor}
\usepackage{caption}
\usepackage{subcaption}
\usepackage{dsfont}

\usepackage{siunitx}

\usepackage[colorinlistoftodos]{todonotes}

\usepackage{amsmath}
\usepackage{amssymb}
\usepackage{amsfonts}
\usepackage{bm}
\newcommand{\R}{\mathbb{R}}
\newcommand{\PP}{\mathds{P}}
\newcommand{\E}{\mathbb{E}}

\DeclareMathOperator*{\argmin}{argmin}

\definecolor{midnightblue}{rgb}{0.1, 0.1, 0.44}
\usepackage{authblk}
\usepackage{footnote}
\usepackage{hyperref}
\hypersetup{
    colorlinks = true,
    urlcolor = blue,
    citecolor = midnightblue,
    linkcolor = midnightblue,
}
\usepackage{xurl}
\usepackage{cleveref}
\usepackage[
    sorting=none
]{biblatex}
\usepackage{titlesec}
\titleformat*{\section}{\large\centering\bfseries}
\titlespacing*{\section}{0pt}{16pt}{8pt}
\titlespacing*{\subsection}{0pt}{16pt}{8pt}
\titleformat*{\subsection}{\centering\bfseries}

\usepackage{blindtext}

\begin{document}

\title{\vspace{-0.75in}Precision Machine Learning
}
\author[1, 2]{Eric J. Michaud\thanks{ericjm@mit.edu}}
\author[1, 2]{Ziming Liu}
\author[1, 2, 3]{Max Tegmark}
\affil[1]{Department of Physics, MIT}
\affil[2]{NSF AI Institute for AI and Fundamental Interactions}
\affil[3]{Center for Brains, Minds and Machines}
\date{October 24, 2022}

\maketitle

\begin{abstract}
    We explore unique considerations involved in fitting ML models to data with very high precision, as is often required for science applications. 
   We empirically compare various function approximation methods and study how they \emph{scale} with increasing parameters and data. We find that neural networks can often outperform classical approximation methods on high-dimensional examples, by auto-discovering and exploiting modular structures therein. However, neural networks trained with common optimizers are less powerful for low-dimensional cases, which motivates us to study the unique properties of neural network loss landscapes and the corresponding optimization challenges that arise in the high precision regime. To address the optimization issue in low dimensions, we develop training tricks which enable us to train neural networks to extremely low loss, close to the limits allowed by numerical precision.

\end{abstract}

\section{Introduction}

Most machine learning practitioners do not need to fit their data with much precision. When applying machine learning to traditional AI tasks such as in computer vision or natural language processing, one typically does not desire to bring training loss all the way down to exactly zero, in part because training loss is just a proxy for some other performance measure like accuracy that one actually cares about, or because there is intrinsic uncertainty which makes perfect prediction impossible, e.g., for language modeling. Accordingly, to save memory and speed up computation, much work has gone into \emph{reducing} the numerical precision used in models without sacrificing model performance much~\cite{pmlr-v37-gupta15, micikevicius2017mixed, kalamkar2019study}. However, modern machine learning methods, and deep neural networks in particular, are now increasingly being applied to science problems, for which being able to fit models \emph{very} precisely to (high-quality) data can be important. Small absolute changes in loss can make a big difference, e.g., for the symbolic regression task of identifying an exact formula from data.

It is therefore timely to consider what, if any, unique considerations arise when attempting to fit ML models very precisely to data, a regime we call \textit{Precision Machine Learning (PML)}. How does pursuit of precision affect choice of method? How does optimization change in the high-precision regime? Do otherwise-obscure properties of model expressivity or optimization come into focus when one cares a great deal about precision? In this paper, we explore these basic questions.

\subsection{Problem Setting}

We study regression in the setting of supervised learning, in particular the task of fitting functions $f: \R^d \rightarrow \R$ to a dataset of $D = \{ (\vec{x}_i, y_i = f(\vec{x}_i) \}_{i=1}^{|D|}$. In this work, we mostly restrict our focus to functions $f$ which are given by symbolic formulas. Such functions are appropriate for our purpose, of studying precision machine learning for science applications, since they (1) are ubiquitous in science, fundamental to many fields' descriptions of nature, (2) are precise, not introducing any intrinsic noise in the data, making extreme precision possible, and (3) often have interesting structure such as \emph{modularity} that sufficiently clever ML methods should be able to discover and exploit. We use a dataset of symbolic formulas from~\cite{udrescu2020ai}, collected from the Feynman Lectures on Physics~\cite{leighton1965feynman}.

Just how closely can we expect to fit models to data? When comparing a model prediction $f_\theta(\vec{x}_i)$ to a data point $y_i$, the smallest nonzero difference allowed is determined by the numerical precision used. IEEE 754 64-bit floats \cite{ieeestandard2019} have 52 mantissa bits, so if $y_i$ and $f_\theta(\vec{x}_i)$ are of order unity, then the smallest nonzero difference between them is $\epsilon_0 = 2^{-52} \sim 10^{-16}$. We should not expect to achieve \emph{relative RMSE loss}  below $10^{-16}$, where relative RMSE loss, on a dataset $D$, is:
\begin{equation}
	\ell_{\text{rms}} \equiv \left( \frac{\sum_{i=1}^{|D|} |f_\theta(\vec{x}_i) - y_i|^2}{\sum_{i=1}^{|D|} y_i^2} \right)^{\frac{1}{2}} = \frac{|f_\theta(\vec{x}_i) - y_i|_{\text{rms}}}{y_{\text{rms}}}.
\end{equation}
 In practice, precision can be bottlenecked earlier by the computations performed within the model $f_\theta$. The task of precision machine learning is to try to push the loss down many orders of magnitude, driving $\ell_{\text{rms}}$ as close as possible to the numerical noise floor $\epsilon_0$.

\subsection{Decomposition of Loss}

One can similarly define \emph{relative MSE loss} $\ell_\text{mse} \equiv \ell_\text{rms}^2$, as well as non-relative (standard) MSE loss $L_\text{mse}(f) = \frac{1}{|D|}\sum_{i=1}^D (f_\theta(\vec{x}_i) - y_i)^2$, and $L_\text{rms} = \sqrt{L_\text{mse}}$.
Minimizing $\ell_\text{rms}, \ell_\text{mse}, L_\text{rms}, L_\text{mse}$ are equivalent up to numerical errors. Note that (relative) expected loss can be defined on a probability distribution $\PP_{(\R^d, \R)}$, like so:
\begin{equation}
    \ell^\PP_\text{rms} = \left(\frac{\mathop{\E}_{(\vec{x}, y) \sim \PP}[(f_\theta(\vec{x})-y)^2]}{\mathop{\E}_{(\vec{x}, y) \sim \PP} [y^2]}\right)^{\frac{1}{2}}.
\end{equation}
When we wish to emphasize the distinction between loss on a dataset $D$ (empirical loss) and a distribution $\PP$ (expected loss), we write $\ell^D$ and $\ell^\PP$. In the spirit of~\cite{guhring2020expressivity}, we find it useful to decompose sources of error into different sources, which we term
\emph{optimization error}, 
\emph{sampling luck}, 
the \emph{generalization gap}, and \emph{architecture error}. 
A given model architecture parametrizes a set of expressible functions $\mathcal{H}$. One can define three functions of interest within $\mathcal{H}$: 
\begin{equation}
    f^\text{best}_\PP \equiv\underset{f \in \mathcal{H}}{\argmin}\{ \ell^\PP(f)\},
\end{equation}
the best model on the expected loss $\ell^\PP$,
\begin{equation}
    f^\text{best}_D \equiv \underset{f \in \mathcal{H}}{\argmin}\{ \ell^D(f) \},
\end{equation}
the best model on the empirical loss $\ell^D$, and
\begin{equation}
    f_D^\text{used} = \mathcal{A}(\mathcal{H}, D, L),
\end{equation}
the model found by a given learning algorithm $\mathcal{A}$ which performs possibly imperfect optimization to minimize empirical loss $L$ on $D$.
We can therefore decompose the empirical loss as follows:
\begin{equation}
    \ell^D(f^\text{used}_D) =
      \underbrace{[\ell^D(f^\text{used}_D) - \ell^D(f^\text{best}_D)]}_\text{optimization error} + 
      \underbrace{[\ell^D(f^\text{best}_D) - \ell^\PP(f^\text{best}_D)]}_\text{sampling luck}
    + \underbrace{[\ell^\PP(f^\text{best}_D) - \ell^\PP(f^\text{best}_\mathbb{P})]}_\text{generalization gap}
    + \underbrace{\ell^\PP(f^\text{best}_\mathbb{P})}_\text{architecture error},
\end{equation}
where all terms are positive except possibly the \emph{sampling luck}, which is zero on average, has a standard deviation shrinking with data size $|D|$ according to the Poisson scaling $|D|^{-1/2}$, and will be ignored in the present paper.
The generalization gap has been extensively studied in prior work, so this paper will focus exclusively on the optimization error and the architecture error. 

To summarize: the architecture error is the best possible performance that a given architecture can achieve on the task, the generalization gap is the difference between the optimal performance on the training set $D$ and the architecture error, and the optimization error is the error introduced by imperfect optimization -- the difference between the error on the training set found by imperfect optimization and the optimal error on the training set. When comparing methods and studying their scaling, it useful to ask which of these error sources dominate. 
We will see that both architecture error and optimization error can be quite important in the high-precision regime, as we will elaborate on in Sections~\ref{sec:linear}-\ref{sec:nonlinear} and \Cref{sec:optimization}, respectively.

\subsection{Importance of Scaling Exponents}\label{sec:intro:scaling}

In this work, one property that we focus on is how methods \emph{scale} as we increase parameters or training data. This builds on a recent body of work on scaling laws in deep learning~\cite{hestness2017deep, kaplan2020scaling, henighan2020scaling, hernandez2021scaling, ghorbani2021scaling, gordon2021data, zhai2022scaling, hoffmann2022training, clark2022unified} which has found that, on many tasks, loss decreases predictably as a power-law in the number of model parameters and amount of training data. Attempting to understand this scaling behavior, \cite{sharma2020neural, bahri2021explaining} argue that in some regimes, cross-entropy and MSE loss should scale as $N^{-\alpha}$, where $\alpha \gtrsim 4/d$, $N$ is the number of model parameters, and $d$ is the \emph{intrinsic dimensionality} of the \emph{data manifold} of the task.

\begin{sloppypar}
Consider the problem of approximating some analytic function ${f : [0, 1]^d \rightarrow \R}$ with some function which is a piecewise $n$-degree polynomial. If one partitions a hypercube in $\R^d$ into regions of length $\epsilon$ and approximates $f$ as a $n$-degree polynomial in each region (requiring $N = \mathcal{O}(1/\epsilon^d)$ parameters), absolute error in each region will be $\mathcal{O}(\epsilon^{n+1})$ (given by the degree-$(n+1)$ term in the Taylor expansion of $f$) and so absolute error scales as $N^{-\frac{n+1}{d}}$.
If neural networks use ReLU activations, they are piecewise linear, $n=1$ and so we may expect $\ell_\text{rmse}(N) \propto N^{-\frac{2}{d}}$. However, in line with~\cite{sharma2020neural}, we find that ReLU NNs often scale as if the problem was lower-dimensional than the input dimension, though we suggest that this is a result of the computational modularity of the problems in our setting, rather than a matter of low intrinsic dimensionality (though these perspectives are related). 
\end{sloppypar}

If one desires very low loss, then the exponent $\alpha$, the rate at which methods approach their best possible performance\footnote{The best possible performance can be determined either by precision limits or by noise intrinsic to the problem, such as intrinsic entropy of natural language.} matters a great deal. Kaplan et al.~\cite{sharma2020neural} note that $4/d$ is merely a lower-bound on the scaling rate -- we consider ways that neural networks can improve on this bound. Understanding model scaling is key to understanding the feasibility of achieving high precision.

\subsection{Organization}

This paper is organized as follows: In Section~\ref{sec:linear} we discuss piecewise linear approximation methods, comparing ReLU networks with linear simplex interpolation. We find that neural networks can sometimes outperform simplex interpolation, and suggest that they do this by discovering modular structure in the data. In Section~\ref{sec:nonlinear} we discuss nonlinear methods, including neural networks with nonlinear activation functions. In Section~\ref{sec:optimization} we discuss the optimization challenges of high-precision neural network training -- how optimization difficulties can often make total error far worse than the limits of what architecture error allows. We attempt to develop optimization methods for overcoming these problems and describe their limitations, then conclude in Section \ref{sec:conclusions}. 

\begin{figure}[t]
    \centering
    \begin{subfigure}{\linewidth}
        \centering
        \includegraphics{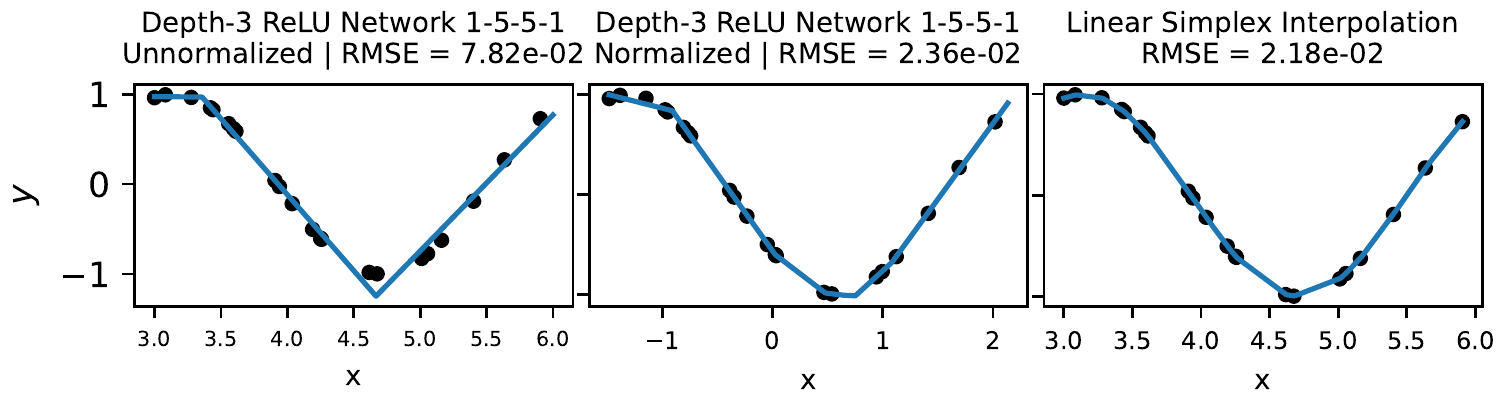}
        \caption{}
        \label{fig:1d-linear-regions}
    \end{subfigure}
    \hfill
    \begin{subfigure}{\linewidth}
        \centering
        \includegraphics{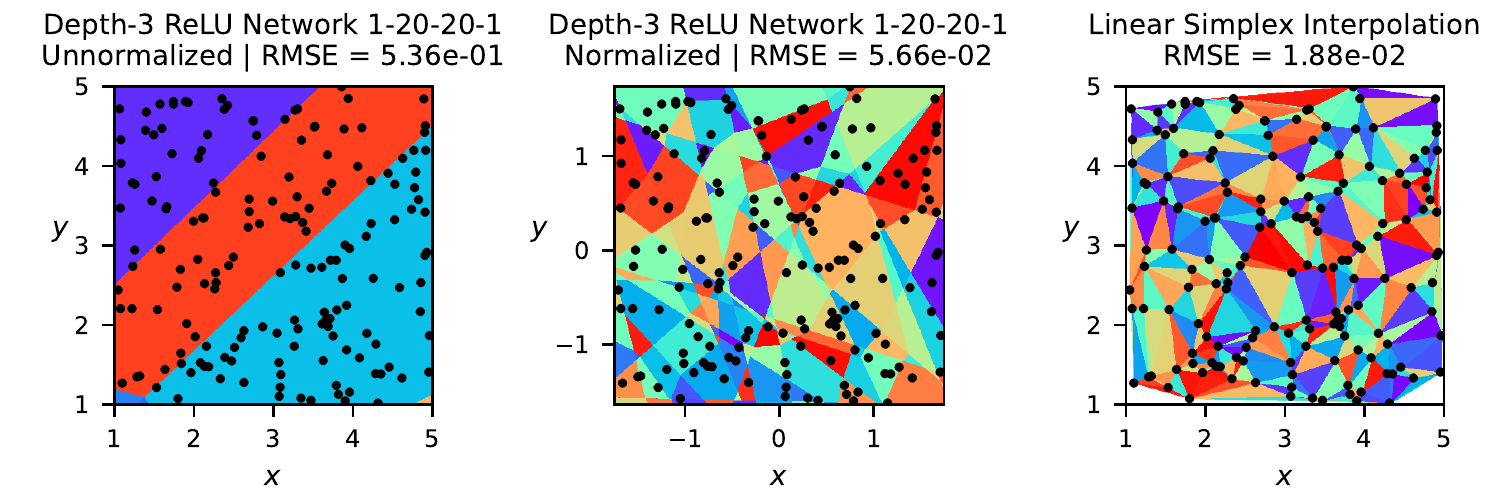}
        \caption{}
        \label{fig:2d-linear-regions}
     \end{subfigure}
    \caption{In \ref{fig:1d-linear-regions} (top), we show the solutions learned by a ReLU network and linear simplex interpolation on the 1D problem $y = \cos(2x)$. In \ref{fig:2d-linear-regions} (bottom), we visualize linear regions for a ReLU network, trained on unnormalized data (left) and normalized data (center), as well as linear simplex interpolation (right) on the 2D problem $z = xy$. In general, we find that normalizing data to have zero mean and unit variance improves network performance, but that linear simplex interpolation outperforms neural networks on low-dimensional problems by better vertex placement.}
    \label{fig:2d-regions}
\end{figure}

\section{Piecewise Linear Methods}
\label{sec:linear}

\begin{figure}[t]
    \centering
    \includegraphics{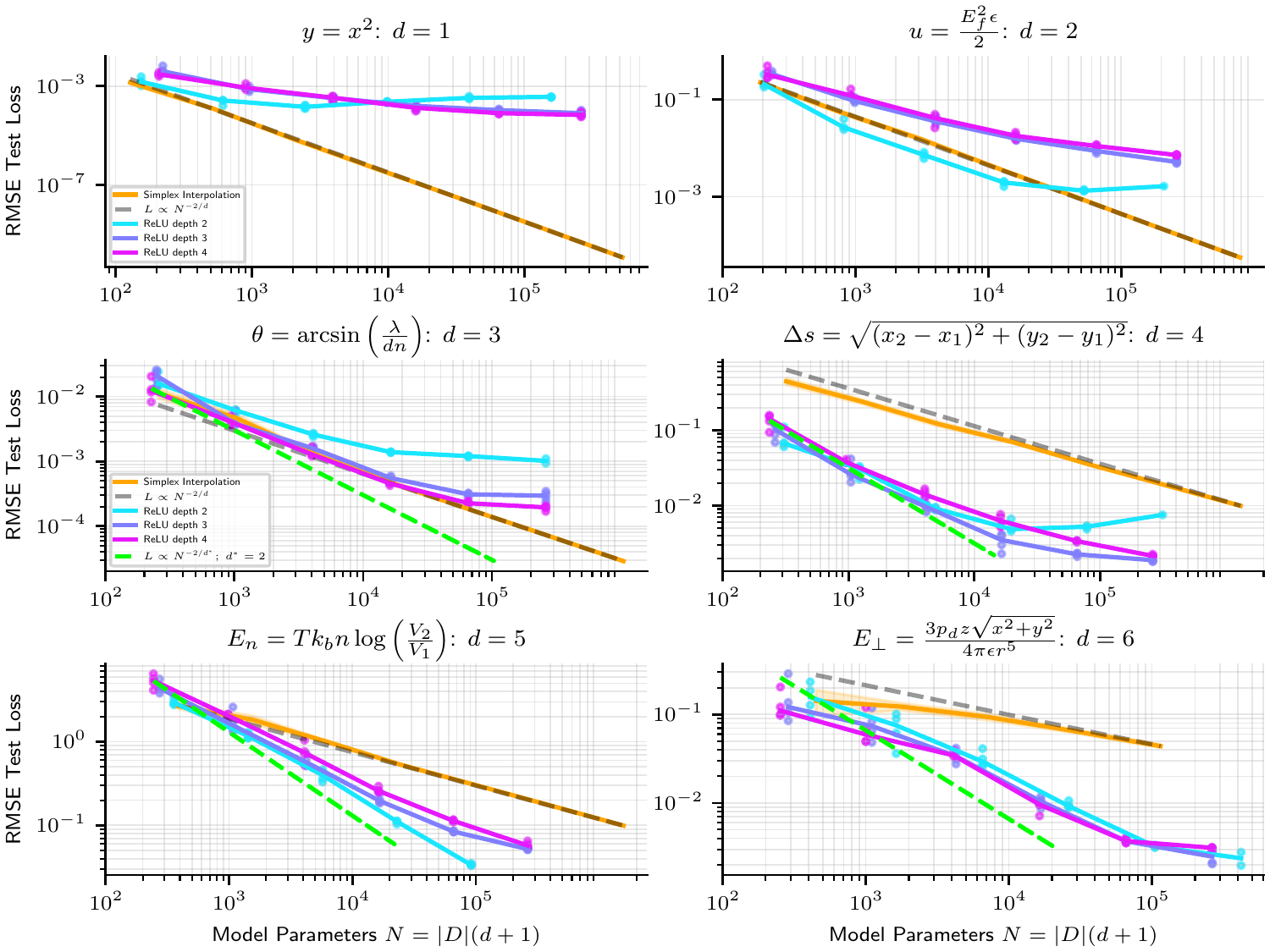}
    \caption{Scaling of linear simplex interpolation versus ReLU NNs. While simplex interpolation scales very predictably as $N^{-2/d}$, where $d$ is the input dimension, we find that NNs sometimes scale better (at least in early regimes) as $N^{-2/{d^*}}$, where $d^* = 2$, on high dimensional problems.}
    \label{fig:linear-scaling-comparison}
\end{figure}

\begin{figure}[t]
    \centering
    \includegraphics{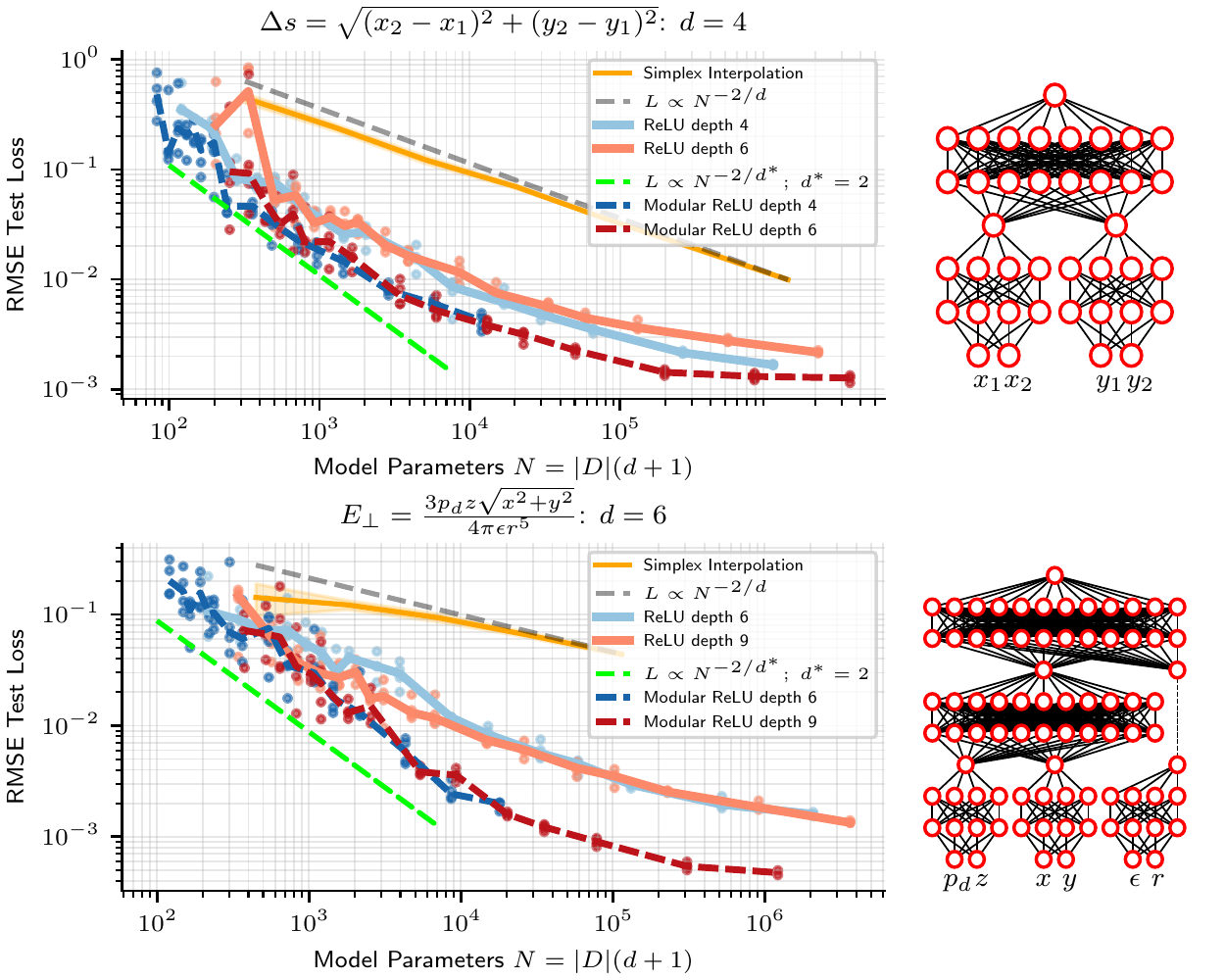}
    \caption{ReLU neural networks are seen to initially scale roughly as if they were modular. Networks with enforced modularity (dark blue and red, dashed line), with architecture depicted on the right, perform and scale similarly, though slightly better, than standard dense MLPs of the same depth (light blue and red).}
    \label{fig:modularity-comparison}
\end{figure}

We first consider approximation methods which provide a piecewise linear fit to data. We focus on two such methods: linear simplex interpolation and neural networks with ReLU activations. 

\begin{sloppypar}
To review, linear simplex interpolation works as follows: given our dataset of $|D|$ input-output pairs $\{(\vec{x}_i, y_i)\}_{i=1}^{|D|}$, linear simplex interpolation first computes a Delaunay triangulation from $\vec{x}_1, \ldots, \vec{x}_{|D|}$ in the input space $\R^d$, partitioning the space into a collection of $d$-simplices,  each with $d+1$ vertices, whose union is the convex hull of the input points. Since $d+1$ points determine a linear (affine) function $\R^d \rightarrow \R$, the function $f$ can be approximated within each $d$-simplex as the unique linear function given by the value of the function $f$ at the vertices. This gives a piecewise linear function on the convex hull of the training points. Linear simplex interpolation needs to store $N = |D|(d+1)$ parameters: $|D|d$ values for the vertices $\vec{x}_i$, and $|D|$ values for the corresponding function values $y_i$.
\end{sloppypar}

Neural networks with ReLU activations also give a piecewise linear fit $f_\theta$. We consider only fully-connected feedforward networks, a.k.a. multilayer perceptrons (MLPs). Such networks consist of a sequence of alternating affine transformations $T : \vec{x} \mapsto W\vec{x} + b$ and element-wise nonlinearities $\sigma(\vec{x})_i = \sigma(\vec{x}_i)$ for an activation function $\sigma : \R \rightarrow \R$:
$$ f_\theta = T_{k+1} \circ \sigma \circ T_k \circ \cdots \circ T_2 \circ \sigma \circ T_1 $$
Following~\cite{arora2016understanding}, we define the depth of the network as the number of affine transformations in the network, which is one greater than the number of hidden layers $k$. As shown in~\cite{arora2016understanding}, any piecewise linear function on $\R^d$ can be represented by a sufficiently wide ReLU NN with at most $\lceil\log_2(d+1)\rceil+1$ depth. Therefore, sufficiently wide and deep networks are able to exactly express functions given by linear simplex interpolation. A natural question then is: given the same amount of data and parameters, how do the two methods compare? We find that simplex interpolation performs better on 1D and 2D problems, but that neural networks can outperform simplex interpolation on higher-dimensional problems. So although simplex interpolation and ReLU NNs both parametrize the same function class (piecewise linear functions), their performance can differ significantly in practice. 

In our experiments, we use the implementation of simplex interpolation from SciPy~\cite{virtanen2020scipy}. When training neural networks, we use the Adam optimizer~\cite{kingma2014adam} with a learning rate of $10^{-3}$, and train for 20k steps. We use a batch size of $\min(|D|, 10^4)$. While we report loss using RMSE, we train using MSE loss. Training points are sampled uniformly from intervals specified by the AI Feynman dataset~\cite{udrescu2020ai} (typically $[1, 5] \subset \R$ for each input), but when training neural networks, we normalize the input points~\cite{lecun2012efficient} so that they have zero mean and unit variance along each dimension. We estimate test loss on datasets of 30k samples.\footnote{Project code can be found at \url{https://github.com/ejmichaud/precision-ml}.} 

In \Cref{fig:2d-regions}, we show for 1D and 2D problems the linear regions given both by simplex interpolation and by neural networks trained a with comparable number of parameters . For 2D problems, \Cref{fig:2d-linear-regions} illustrates the importance of normalizing input data for ReLU networks. We see that there is a far higher density of linear regions around the data when input data is normalized, which leads to better performance. 
Neural networks, with the same number of parameters and trained with the same low-dimensional data, often have fewer linear regions than simplex interpolation.

In Figure~\ref{fig:linear-scaling-comparison}, we show how the precision of linear simplex interpolation and neural networks scale empirically. Since simplex interpolation is a piecewise linear method, from the discussion in~\Cref{sec:intro:scaling}, we expect its RMSE error to scale as $N^{-2/d}$, and find that this indeed holds\footnote{Scaling as $D^{-2/d}$ only holds when the model is evaluated on points not too close to the boundary of the training set. At the boundary, simplices are sometimes quite large, leading to a poor approximation of the target function close to the boundary, large errors, and worse scaling. In our experiments, we therefore compute test error only for points at least 10\% (of the width of the training set in each dimension) from the boundary of the training set.}. To provide a fair comparison with simplex interpolation when evaluating neural networks on a dataset of size $D$, 
we give it the same number of parameters
$N = |D|(d+1)$. From~\Cref{fig:linear-scaling-comparison}, we see that simplex interpolation outperforms neural networks on low dimensional problems but that neural networks do better on higher-dimensional problems.

For the 1D example in ~\Cref{fig:linear-scaling-comparison} (top left), we know that the amount by which the neural networks under-perform simplex interpolation is entirely due to optimization error. This is because any 1D piecewise linear function $f(x)$ with $m$ corners at $x_1,...,x_m$ can trivially be can trivially be written as a linear combination of $m$ functions ReLU$(x-x_i)$. 

Interestingly, we see that, at least early in the scaling curves, neural networks usually appear to scale not as $N^{-2/d}$, but rather as $N^{-2/d^*}$ where $d^*$ is the \emph{maximum arity} of the problem. Symbolic expressions typically have modular structure, and can be viewed as a series of computations each acting on fewer variables than are in the whole expression. For instance, for the expression $x_1\cdot x_2 \cdot x_3$, one can decompose the operation as a multiplication between $x_1$ and $x_2$, and then a second multiplication between the result and $x_3$. At each stage, only two variables are operated on, so the maximum arity of the computational graph is $2$. If neural networks can discover this modularity, they can scale as $N^{-2/d^*}$ where $d^* = 2$, and outperform simplex interpolation when $d > 2$.

To test this idea that better neural network scaling comes from exploiting modularity, we train networks where we hard-code the modularity of the problem into the architecture, as depicted in~\Cref{fig:modularity-comparison}. Figure~\ref{fig:modularity-comparison} indeed reveals how models for which we enforce the modularity of the problem perform and scale similarly to same-depth dense neural networks without modularity enforced. A modular architecture can be created from a dense one by forcing weight matrices to be block-diagonal (where we do not count off-diagonal entries towards the number of model parameters), but in practice we create modular architectures by creating a separate MLP for each node in the symbolic expression computation graph and connecting them together in accordance with the computation graph. See the diagrams in Figure~\ref{fig:modularity-comparison} for an illustration of the modular architecture. In Figure~\ref{fig:modularity-comparison}, we plot modular and dense network performance against number of model parameters, but we also find that holding width constant, rather than number of parameters, modular networks still slightly outperform their dense counterparts. For instance, depth-6 width-100 modular networks outperform dense networks of the same width and depth, despite dense networks having $\approx$2.5x fewer parameters.
Such ``less is more'' results are to be expected if the optimal architecture is in fact modular, in which case a fully connected architecture wastes resources training large numbers of parameters that should be zero.

The fact that neural networks can scale as if the problem dimension was the maximum arity of the computational graph, rather than the input dimension, is similar to, although more general than, a result from~\cite{sharma2020neural}. They found that if the problem ``data manifold'' is a product $X_1 \times X_2 \times \cdots \times X_n$, and the prediction problem decomposes as $F(x) = \sum_i f_i(x_i)$, then the effective dimension is given by the maximum dimension of the manifolds $X_1, \ldots, X_n$. Successfully scaling in the maximum arity of the computation graph requires the network to learn a particular \emph{compositional} structure consisting of modules which are sparse, acting on a lower number of variables. This relates to a literature beginning to emerge on compositional sparsity in deep learning~\cite{dahmen2022compositional, poggio2017and}.

\section{Nonlinear Methods}
\label{sec:nonlinear}

\begin{figure}
    \centering
    \includegraphics{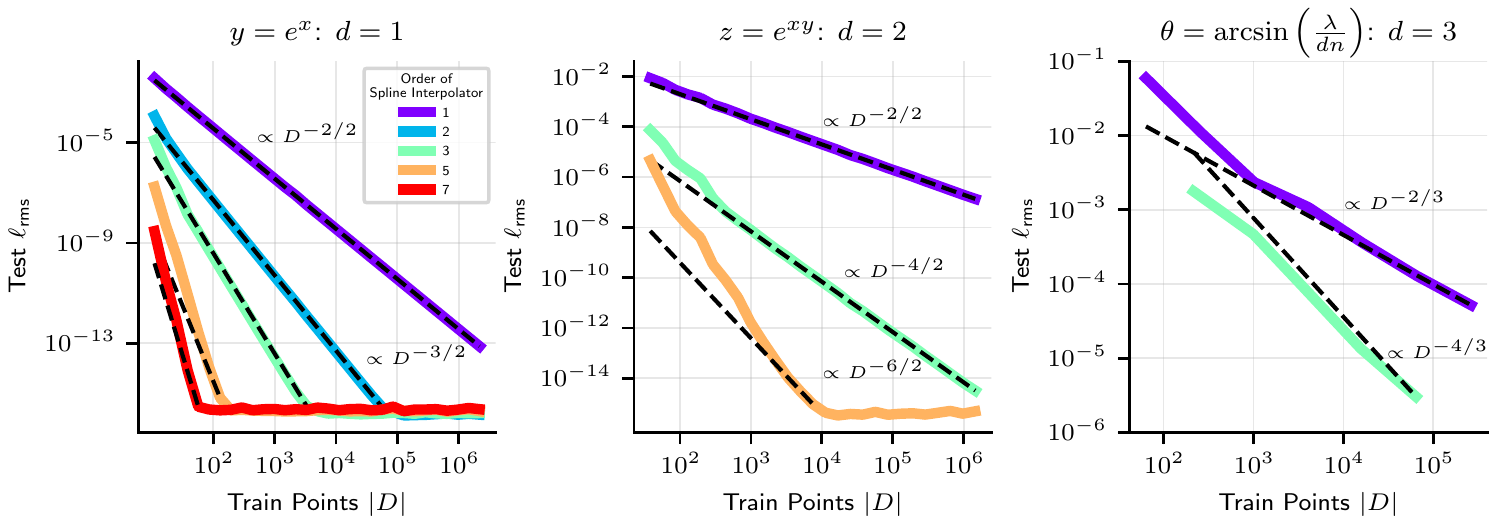}
    \caption{Interpolation methods, both linear and nonlinear, on 2D and 3D problems, seen to approximately scale as $D^{-(n+1)/d}$ where $n$ is the order of the polynomial spline, $d$ is the input dimension.}
    \label{fig:2d3dmethods}
\end{figure}

We now turn our attention to approximations methods that are thoroughly nonlinear (as opposed to piecewise linear).
As discussed in the introduction, methods approximating the target function $f$ by a piecewise polynomial have a scaling exponent $\alpha = \frac{n+1}{d}$ where $n$ is the degree of the polynomial. 

In Figure~\ref{fig:2d3dmethods}, we plot the performance of approximation methods which are piecewise polynomial, for 1D, 2D and 3D problems. For 1D and 2D problems, we use splines of varying order. For 3D problems, we use the cubic spline interpolation method of ~\cite{lekien2005tricubic}. We see empirically that these methods have scaling exponent $\alpha = (n+1)/d$. If the order of the spline interpolator is high enough, and the dimension low enough, we see that relative RMSE loss levels out at $\epsilon_0 \approx 10^{-16}$ at the precision limit. Unfortunately, a basic limitation of these methods is that they are limited to low-dimensional problems. 

\begin{figure}
    \centering
    \includegraphics{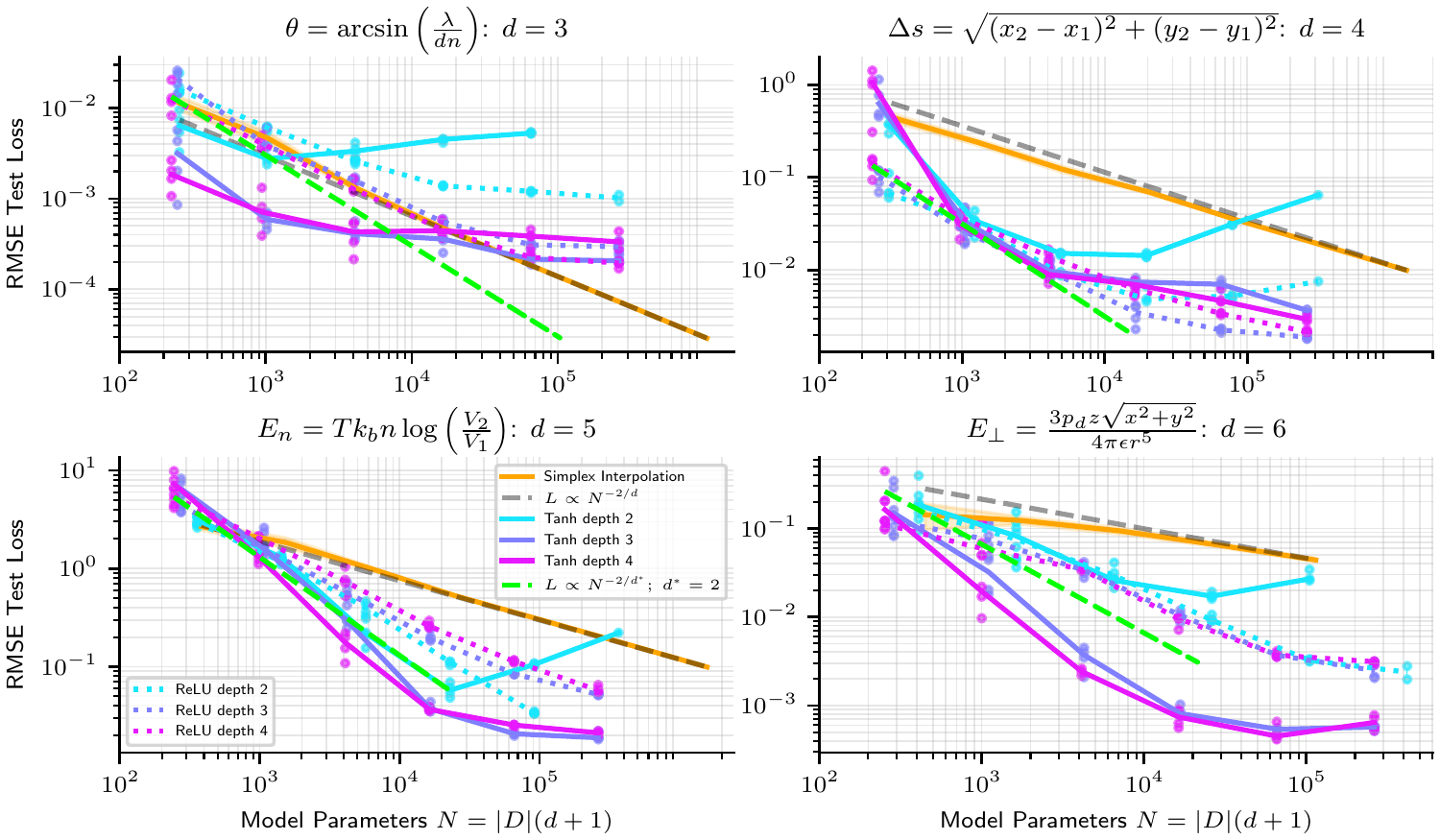}
    \caption{Scaling of linear simplex interpolation vs tanh NNs. We also plot ReLU NN performance as a dotted line for comparison. While simplex interpolation scales very predictably as $N^{-2/d}$, where $d$ is the input dimension, tanh NN scaling is much messier.}
    \label{fig:tanh-scaling-comparison}
\end{figure}

There are relatively few options for high-dimensional nonlinear interpolation.\footnote{One method, which we have not tested, is radial basis function interpolation.} A particularly interesting parametrization of nonlinear functions is given by neural networks with nonlinear activation functions (and not piecewise linear like ReLUs). In Figure~\ref{fig:tanh-scaling-comparison}, we show how neural networks with tanh activations scale in increasing width. We observe that on some problems, they do better than the ideal scaling achievable with linear methods (shown as a green dashed line). However, in our experiments, they can sometimes scale worse, perhaps the result of imperfect optimization. Also, we find that scaling is typically not nearly as clean as a power law as it was for ReLU networks. 

\begin{figure}
     \centering
     \begin{subfigure}{0.65\textwidth}
         \centering
         \includegraphics[height=2in]{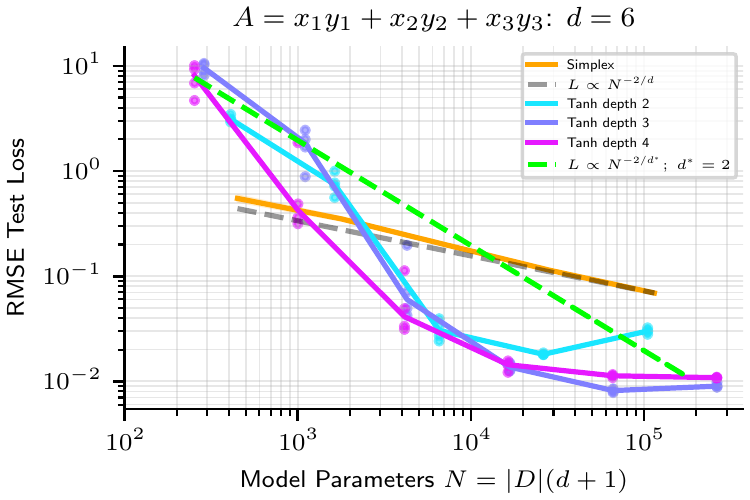}
         \caption{}
         \label{fig:exactly-expressible-subfigure}
     \end{subfigure}
     \hfill
     \begin{subfigure}{0.3\textwidth}
         \centering
         \includegraphics[height=2in]{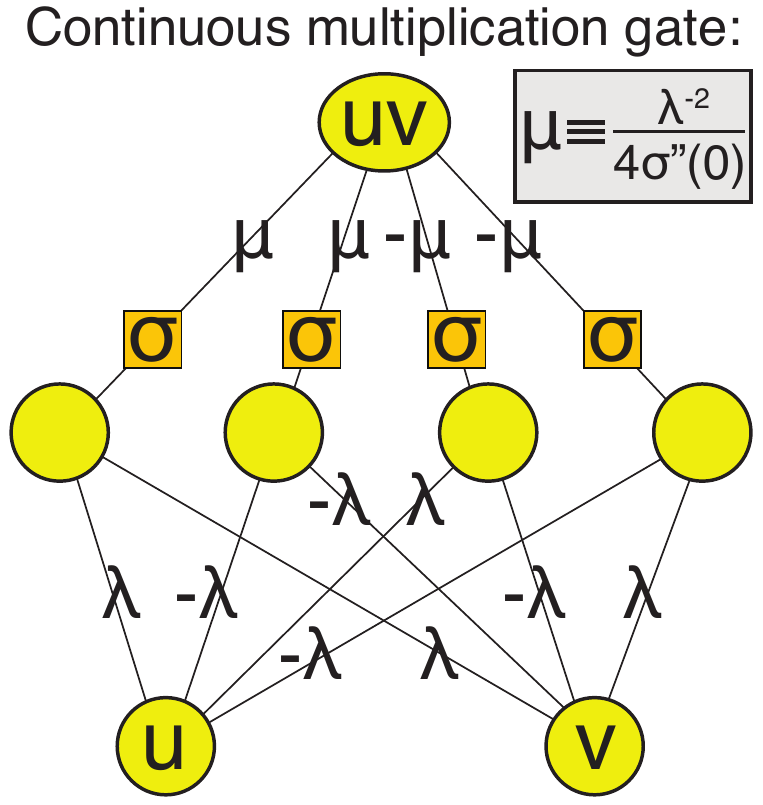}
         \caption{}
         \label{fig:multiplication-diagram}
     \end{subfigure}
    \caption{(a) Scaling of neural networks on a target function which can be arbitrarily closely approximated by a network of finite width. (b) diagram from~\cite{lin2017does} showing how a 4-neuron network can implement multiplication arbitrarily well. Therefore a depth-2 network of width at least 12 has an architecture error at the machine precision limit, yet optimization in practice does not discover solutions within at least 10 orders of magnitude of the precision limit.}
\end{figure}

For some problems, one can show theoretically that architecture error can be made arbitrarily low, and that the loss is due entirely to optimization error and the generalization gap. As shown in~\cite{lin2017does}, a two-layer neural network with only four hidden units can perform multiplication between two real numbers, provided that a twice-differentiable activation function is used. See \Cref{fig:multiplication-diagram} for a diagram of such a network, taken from~\cite{lin2017does}. Note that this network becomes more accurate in the limit that some of its parameters become very small and others become very large. This result, that small neural networks can express multiplication arbitrarily well, implies that neural network architecture error is effectively zero for some problems. However, actually \emph{learning} this multiplication circuit in practice is challenging since it involves some network parameters \emph{diverging} $\rightarrow \infty$ while others $\rightarrow 0$ in a precise ratio. This means that for some tasks, neural network performance is mainly limited not by architecture error, but by optimization error.

Indeed, on some problems, a failure to achieve high precision can be blamed entirely on the optimization error. In \Cref{fig:exactly-expressible-subfigure}, we show neural network scaling on the equation $f(x_1, x_2, x_3, y_1, y_2, y_3) = x_1y_1 + x_2y_2 + x_3y_3$. For this problem, a 2-layer network with 12 hidden units (implementing three multiplications in parallel, with their results added in the last layer) can achieve $\approx 0$ architecture error. Yet we see a failure to get anywhere near that architecture error or the noise floor set by machine precision. Instead, as one scales up the network size on this task, we see that despite architecture error abruptly dropping to zero early on, the actually attained loss continues to scales down smoothly. 

It is therefore important to analyze the problem of optimization for high precision, which we do in the next section.

\section{Optimization}
\label{sec:optimization}

As seen above, when deep neural networks are trained with standard optimizers, they can produce significant optimization error, i.e. fail to find the best approximation. In this section, we discuss the difficulty of optimization in the high-precision regime and explore a few tricks for improving neural network training.

\subsection{Properties of Loss Landscape}

\begin{figure}[ht]
    \centering
    \includegraphics[width=4in]{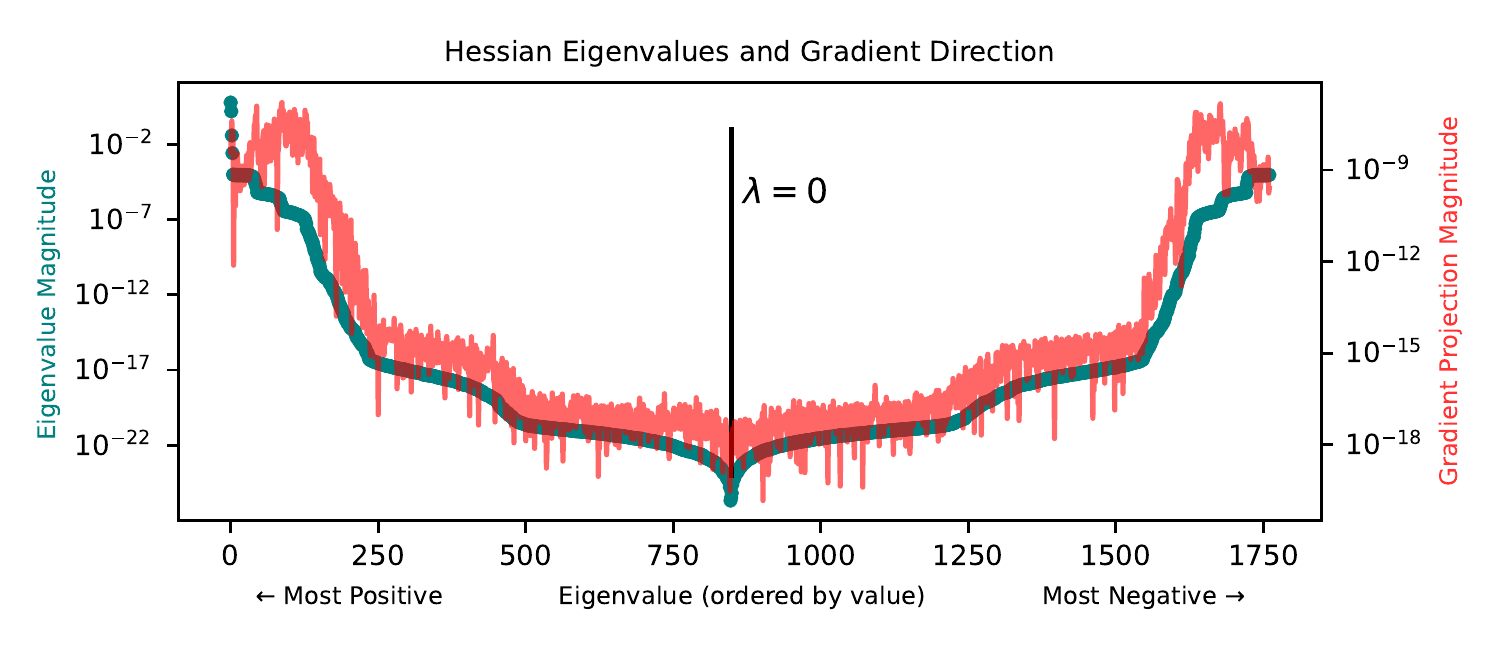}
    \caption{Eigenvalues (dark green) of the loss landscape Hessian (MSE loss) after training with the Adam optimizer, along with the magnitude of the gradient's projection onto each corresponding eigenvector (thin red line). We see a cluster of top eigenvalues and a bulk of near-zero eigenvalues. The gradient  (thin jagged red curve) points mostly in directions of high-curvature.}
    \label{fig:eigenvalue-spectrum}
\end{figure}

To understand the difficulty of optimizing in the high-precision regime, we attempt to understand the local geometry of the loss landscape at low loss. In particular, we compute the Hessian of the loss and study its eigenvalues. In \Cref{fig:eigenvalue-spectrum}, we plot the spectrum of the Hessian, along with the magnitude of the gradient projected onto each of the corresponding eigenvectors, at a point in the loss landscape found by training with the Adam optimizer for 30k steps in a teacher-student setup. The teacher is a depth-3, width-3 tanh MLP and the student is a depth-3, width-40 tanh MLP. In line with~\cite{sagun2016eigenvalues, sagun2017empirical, gur2018gradient}, we find that at low loss, the loss landscape has a top cluster of directions of high curvature (relatively large positive eigenvalues) and a bulk of directions of very low curvature. Furthermore, the gradient tends to point most strongly in directions of higher curvature, and has very little projection onto directions of low curvature magnitude.

The basic picture emerging from this analysis is that of a canyon, i.e., a very narrow, very long valley around a low-loss minimum. The valley has steep walls in high-curvature directions and a long basin in low-curvature directions. Further reducing loss in this environment requires either (1) taking very precisely-sized steps along high-curvature directions to find the exact middle of the canyon or (2) moving along the canyon in low-curvature directions instead, almost orthogonally to the gradient. In this landscape, typical first-order optimizers used in deep learning may struggle to do either of these things, except perhaps if learning rates are chosen extremely carefully. Instead, they tend to waste time rapidly oscillating between the side walls of the canyon.

\subsection{Optimization Tricks For Reducing Optimization Error}

How can we successfully optimize in such a poorly-conditioned, low-loss regime? We first find that switching from first-order optimizers like Adam to second-order optimizers like BFGS~\cite{nocedal1999numerical} can improve RMSE loss by multiple orders of magnitude. Second-order methods often both (1) employ line searches, and (2) search in directions not strongly aligned with the gradient, allowing optimization to progress within low-curvature subspaces. However, methods like BFGS are eventually bottlenecked by numerical precision limitations. To further lower loss, we tested the following two methods:

\paragraph{Low-Curvature Subspace Optimization} We find that by restricting our optimization to low-curvature subspaces, we can further decrease loss past the point where loss of precision prevented BFGS from taking further steps. Our method has a single hyperparameter $\tau$. The method is as follows: let $g = \nabla_\theta \mathcal{L}$ be the gradient and $H$ be the Hessian of the loss. Denote an eigenvector-eigenvalue pair of $H$ by $(e_i, \lambda_i)$. Instead of stepping in the direction $-g$, we instead compute $\hat{g} = \sum_{i : \lambda_i < \tau} e_i (e_i \cdot g)$. Essentially, we just project $g$ onto the subspace spanned by eigenvalues of $e_i$ such that $\lambda_i < \tau$. We then perform a line search to minimize loss along the direction $-\hat{g}$, and repeat the process. Note that this requires computing eigenvectors and eigenvalues for the whole Hessian $H$.

\begin{figure}[t]
    \centering
    \includegraphics{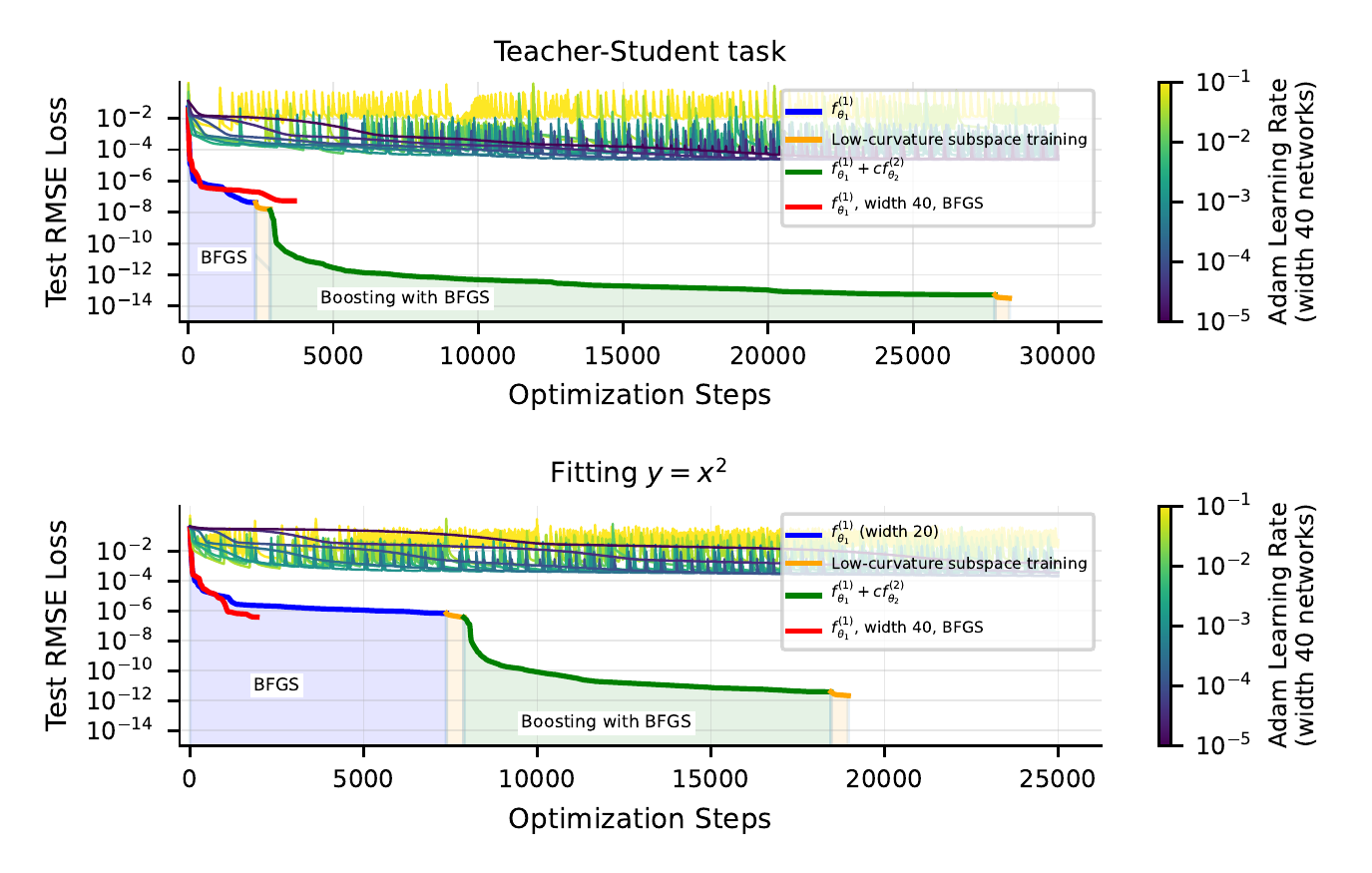}
    \caption{Comparison of Adam with BFGS + low-curvature subspace training + boosting. Using second-order methods like BFGS, but especially using boosting, leads to an improvement of many orders of magnitude over just training with Adam. Target functions are a teacher network (\textbf{top}) and a symbolic equation (\textbf{bottom}).
    }
    \label{fig:combined-techniques-learning-curves}
\end{figure}

\paragraph{Boosting: staged training of neural networks} Instead of training a full network to fit the target $f$, one can train two networks $f^{(1)}_{\theta_1}, f^{(2)}_{\theta_2}$ sequentially: first train $f^{(1)}_{\theta_1}$ to fit $f$, then train $f^{(2)}_{\theta_2}$ to fit the residual $\frac{f - f^{(1)}_{\theta_1}}{c}$, where $c \ll 1$ normalizes the residual to be of order unity. One can then combine the two networks into a single model $f(x) \approx f^{(1)}_{\theta_1}(x) + c f^{(2)}_{\theta_2}(x)$. If networks $f^{(1)}_{\theta_1}, f^{(2)}_{\theta_2}$ have widths $w_1, w_2$ respectively, then they can be combined into one network of width $w_1, w_2$, with block-diagonal weight matrices, and where the parameters of the last layer of $f^{(2)}_{\theta_2}$ are scaled down by $c$.

We find that, for low-dimensional problems, we can achieve substantially lower loss with these techniques. We use the following setup: we train width-40 depth-3 tanh MLPs to fit single-variable polynomials with the BFGS optimizer on MSE loss. The SciPy~\cite{virtanen2020scipy} BFGS implementation achieves $~10^{-7}$ RMSE loss before precision loss prevents further iterations. Subsequently using low-curvature subspace training with a threshold $\tau = 10^{-16}$ can further lower RMSE loss a factor of over 2x. On similar low-dimensional problems, as shown in Figure~\ref{fig:combined-techniques-learning-curves}, applying boosting, training a second network with BFGS on the residual of the first can lower RMSE loss further by 5-6 orders of magnitude. In \Cref{fig:combined-techniques-learning-curves}, we compare training runs with these tricks to runs with the Adam optimizer for a variety of learning rates. For our Adam training runs, we use width-40 tanh MLPs. When training with boosting, we train a width-20 network for $f^{(1)}_{\theta_1}$ and a width-20 network for $f^{(2)}_{\theta_2}$, for a combined width of 40. We also plot a width-40 network trained solely with BFGS for comparison. We find, unsurprisingly, that BFGS significantly outperforms Adam. With our tricks, particularly boosting, we can sometimes outperform even well-tuned Adam by 8 orders of magnitude, driving RMSE loss down to $\approx 10^{-14}$, close to the machine precision limit. 

\begin{figure}[t]
    \centering
    \includegraphics{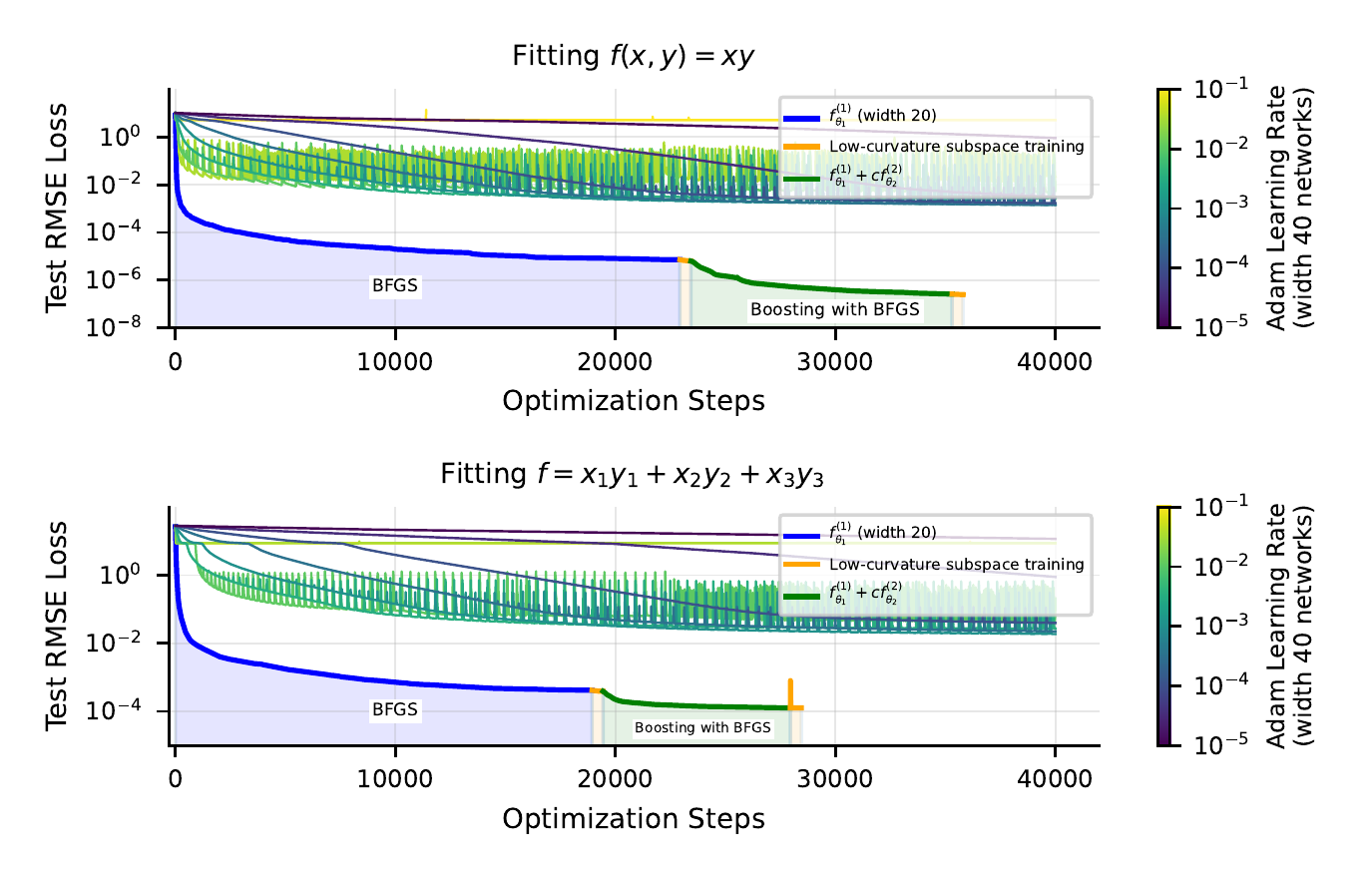}
    \caption{Comparison of Adam with BFGS + low-curvature subspace training + boosting, for a 2D problem (\textbf{top}) and a 6D problem (\textbf{bottom}), the equation we studied in \Cref{fig:exactly-expressible-subfigure}. As we increase dimension, the optimization tricks we tried in this work show diminishing benefits.}
    \label{fig:combined-techniques-learning-curves-6d}
\end{figure}

\subsection{Limitations and Outlook}

The techniques we described above are not a silver bullet for fitting neural networks to any data with high precision. Firstly, second-order optimizers like BFGS scale poorly with the number of model parameters $N$ (since the Hessian is an $N\times N$ matrix), limiting their applicability to small models. Also, we find that the gains from boosting diminish quickly as the input dimension of the problem grows. In \Cref{fig:combined-techniques-learning-curves-6d}, we see that on the 6-dimensional problem discussed earlier (\Cref{fig:exactly-expressible-subfigure}), BFGS + boosting achieves only about a 2-order of magnitude improvement, bringing the RMSE loss from $10^{-2}$ to $10^{-4}$. 

While boosting does not help much for high-dimensional problems, its success on low-dimensional problems is still noteworthy. By training two parts of a neural network separately and sequentially, we were able to dramatically improve performance. This suggests that perhaps there are other methods, not yet explored, for training and assembling neural networks in nonstandard ways to achieve dramatically better precision. The solutions found with boosting, where some network weights are at a much smaller scale than others, are not likely to be found with typical training. An interesting avenue for future work would be exploring new initialization schemes, or other ways of training networks sequentially, to discover better solutions in underexplored regions of parameter space.

\section{Conclusions}
\label{sec:conclusions}

We have studied the problem of fitting scientific data with a variety of approximation methods, analyzing sources of error and their scaling.
\begin{itemize}
\item {\bf Linear Simplex Interpolation} provides a piecewise linear fit to data, with RMSE loss scaling reliably as $D^{-2/d}$. Linear simplex interpolation always fits the training points exactly, and so error comes from the generalization gap and the architecture error:

\item {\bf ReLU Neural Networks} also provide a piecewise linear fit to data. Their performance (RMSE loss) often scales as $D^{-2/{d^*}}$, where $d^*$ is the \emph{maximum arity} of the task (typically $d^* = 2$). Accordingly, they can scale better than linear simplex interpolation when $d > 2$. Unfortunately, they are often afflicted by optimization error making them scale worse than linear simplex interpolation on 1D and 2D problems, and even in higher dimensions in the large-network limit.

\item {\bf Nonlinear Splines}
approximate a target function piecewise by polynomials. They scale as $D^{-(n+1) / d}$ where $n$ is the order of the polynomial.

\item {\bf Neural Networks with smooth activations} provide a nonlinear fit to data. Quite small networks with twice-differentiable nonlinearities can perform multiplication arbitrarily well~\cite{lin2017does}, and so for many of the tasks we study (given by symbolic formulas), their architecture error is zero. We find that their inaccuraccy does not appear to scale cleanly as power-laws. Optimization error is unfortunately a key driver of the error of these methods, but with special training tricks, we found that we could reduce RMSE loss on 1D problems down within 2-4 orders of magnitude of the 64-bit machine precision limit $\epsilon_0 \sim 10^{-16}$.
\end{itemize}

\begin{figure}[ht]
    \centering
    \includegraphics[width=6in]{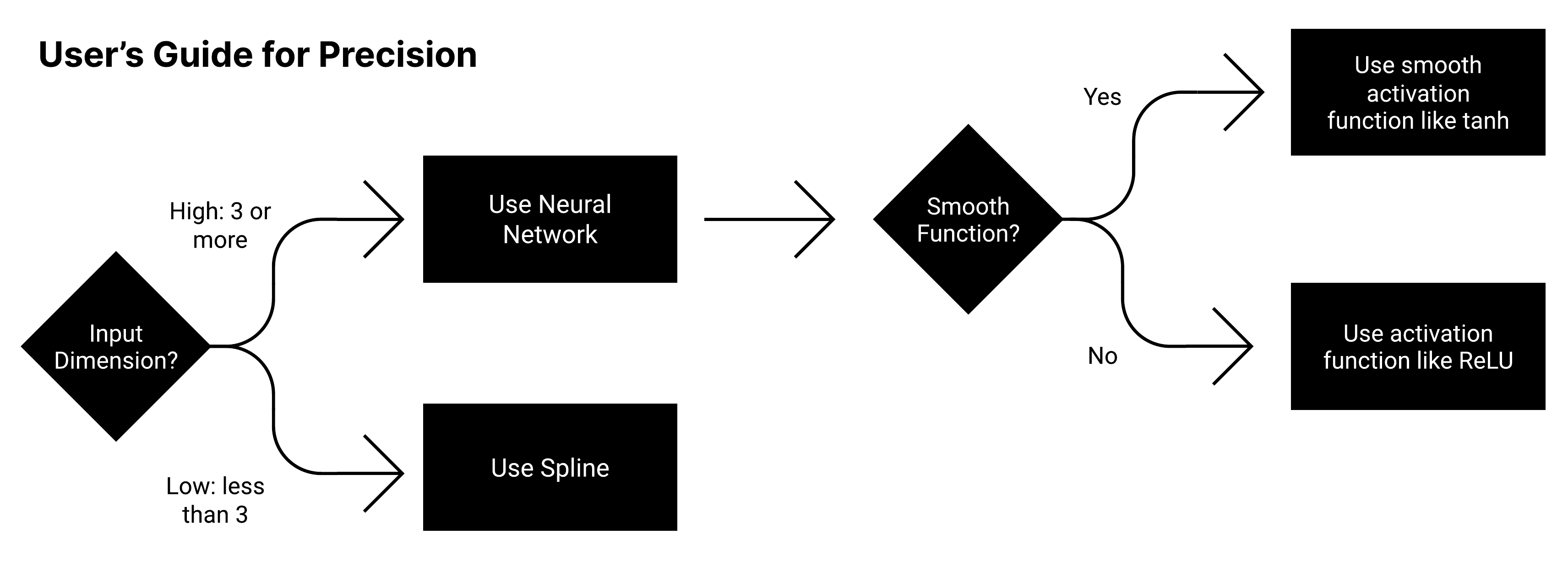}
    \caption{User's Guide for Precision: which approximation is best depends on properties of the problem.}
    \label{fig:flowchart}
\end{figure}

For those seeking high-precision fits, 
These results suggest the following heuristics, summarized in Figure~\ref{fig:flowchart} as a ``User's Guide to Precision": 
If data dimensionality $d$ is low($d \leq 2$), polynomial spline interpolation can provide a fit at machine precision if you (1) have enough data and (2) choose a high enough polynomial order. Neural networks with smooth activations may in some cases also approach machine precision, possibly with less data, if they are trained with second-order optimizers like BFGS and boosted. For higher-dimensional problems ($d \geq 3$), neural networks are typically the most promising choice, since they can learn compositional modular structure that allows them to scale \emph{as if} the data dimensionality were lower.

In summary, our results highlight both advantages and disadvantages of using neural networks to fit scientific data. We hope that they will help provide useful building blocks for further work towards precision machine learning.

{\bf Acknowledgements} 
This work was supported by The Casey
Family Foundation, the Foundational Questions Institute, the Rothberg Family Fund for Cognitive Science, the NSF Graduate Research Fellowship (Grant No. 2141064),
and IAIFI through NSF grant PHY-2019786.

\printbibliography

\end{document}